\documentclass{article}

\usepackage{microtype}
\usepackage{graphicx}
\usepackage{subfigure}
\usepackage{booktabs} 
\usepackage{url}            
\usepackage{booktabs}       
\usepackage{amsfonts}       
\usepackage{nicefrac}       
\usepackage{microtype}      
\usepackage{multirow}
\usepackage[table]{xcolor}
\usepackage{natbib}
\usepackage{adjustbox}

\usepackage{amsmath,amsfonts}
\usepackage{hyperref}
\usepackage{cleveref}

\DeclareMathOperator*{\argmax}{arg\,max}

\newcommand{\eg}{\textit{e.g\@.}}
\newcommand{\etc}{\textit{etc.}}
\newcommand{\etal}{\textit{et~al\@.}}

\usepackage[accepted]{mlsys2020}

\mlsystitlerunning{Submission and Formatting Instructions for MLSys 2020}

\begin{document}

\twocolumn[
\mlsystitle{Learned Low Precision Graph Neural Networks}



\mlsyssetsymbol{equal}{*}

\begin{mlsysauthorlist}
\mlsysauthor{Yiren Zhao}{equal,to}
\mlsysauthor{Duo Wang}{equal,to}
\mlsysauthor{Daniel Bates}{to}
\mlsysauthor{Robert Mullins}{to}
\mlsysauthor{Mateja Jamnik}{to}
\mlsysauthor{Pietro Lio}{to}
\end{mlsysauthorlist}

\mlsysaffiliation{to}{
Department of Compute Science and Technology, University of Cambridge Cambridge, UK}

\mlsyscorrespondingauthor{Yiren Zhao}{yiren.zhao@cl.cam.ac.uk}

\mlsyskeywords{Graph Neural Networks, Quantisation, Network Architecture Search}

\vskip 0.3in

\begin{abstract}
Deep Graph Neural Networks (GNNs) show promising performance on a range of graph tasks, yet at present are costly to run and lack many of the optimisations applied to DNNs.
We show, for the first time, how to systematically
quantise GNNs with minimal or no loss in performance
using Network Architecture Search (NAS).
We define the possible
quantisation search space of GNNs.
The proposed novel NAS mechanism,
named Low Precision Graph NAS (LPGNAS),
constrains both architecture and quantisation
choices to be differentiable.
LPGNAS learns the optimal
architecture coupled with
the best quantisation strategy
for different components in the GNN automatically
using back-propagation in a single search round.
On eight different datasets,
solving the task of classifying unseen nodes in a graph,
LPGNAS generates quantised models with
significant reductions in both model
and buffer sizes but with similar accuracy
to manually designed networks
and other NAS results.
In particular, on the Pubmed dataset,
LPGNAS shows a better size-accuracy Pareto frontier
compared to seven other manual and searched
baselines, offering a $2.3 \times$ reduction in model size but a $0.4\%$ increase in accuracy when compared to the best NAS competitor.
Finally, from our collected quantisation statistics on a wide range of datasets, we suggest a W4A8 (4-bit weights, 8-bit activations) quantisation strategy might be the bottleneck for naive GNN quantisations.
\end{abstract}
]



\printAffiliationsAndNotice{\mlsysEqualContribution} 

\section{Introduction}
Graph Neural Networks (GNNs) have been successful
in fields such as
computational biology \cite{zitnik2017predicting},
social networks \cite{hamilton2017inductive},
knowledge graphs \cite{lin2015learning}, \etc.
The ability of GNNs
to apply learned embeddings to unseen nodes or new subgraphs
is useful for large scale machine learning systems
on graph data, ranging from
analysing posts on forums to
creating product listings for
shopping sites \cite{hamilton2017inductive}.
Most of the large production systems have high throughputs,
running millions or billions of
GNN inferences per second.
This creates an urgent need
to minimise
both the computation and memory cost of
GNN inference.

One common approach to reduce computation
and memory overheads of Deep Neural Networks (DNNs)
is quantisation
\cite{zhao2019focused, zhanglq2018, hubara2016binarized, zhu2016trained, hwang2014fixed}.
A simpler data format not only reduces
the model size but also introduces the chance
of using simpler arithmetic operations on existing
or emerging hardware platforms \cite{zhao2019focused, hubara2016binarized}.
Previous quantisation 
methods focusing
solely on DNNs with image and sequence data
will not obviously transfer well to GNNs.
First,
in CNNs and RNNs it is the convolutional and fully-connected layers that are quantised \cite{hubara2016binarized, shen2019q}.
GNNs, however, follow a sample and aggregate approach
\cite{zhao2020probabilistic, hamilton2017inductive}, where we can separate
a basic building block in GNNs into four smaller sub-blocks
(Linear, Attention, Aggregation and Activation);
only a subset of these sub-blocks
has trainable parameters
and these parameters naturally have different
precision requirements.
Second, the design of GNN blocks involves a different design space,
where several attention and aggregation methods
are available and the choices of these methods
have a direct interaction with the precision.

In this paper, we
propose Low Precision Graph NAS (LPGNAS),
which aims to automatically design quantised GNNs.
The proposed NAS method is single-path,
one-shot and gradient-based.
Additionally, LPGNAS's quantisation options are
at a micro-architecture level so that different sub-blocks
in a graph block can have different quantisation strategies.
In this paper, we make the following contributions.
\begin{itemize}
    \item To our knowledge, this is the first piece of work
    studying how to systematically quantise GNNs.
    We identify the quantisable components and possible search space for GNN quantisation.
    \item We present a novel single-path, one-shot and gradient-based NAS algorithm (LPGNAS)
    for automatically finding low precision GNNs.
    \item We report LPGNAS results,
    and show how they significantly outperfrom
    previous state-of-the-art NAS methods and manually designed networks
    in terms of memory consumption on the same accuracy budget on 8 different datasets.
    \item By running LPGNAS on a wide range of datasets, we reveal the quantisation statistics and show empirically LPGNAS converges to a 4-bit weights 8-bit activation fixed-point quantisation strategy. We further show this quantisation is the limitation of current fixed-point quantisation strategies on manually designed and searched GNNs.
\end{itemize}

\section{Related Work}
\subsection{Quantisation and Network Architecture Search}
DNNs offer great performance and rapid time-to-market path on a broad variety of tasks. Unfortunately, their high memory and computation requirements can be challenging when deploying in real-world scenarios.
Quantisation directly shrinks model sizes
and rapidly simplifies the complexity of the arithmetic hardware.
Previous research shows that DNN models
can be quantised to surprisingly low precisions
such as ternary \cite{zhu2016trained}
and binary \cite{hubara2016binarized}.
Networks with
extremely low precision weights
have a significant task accuracy drop due to the
numerical errors introduced, and sometimes require architectural changes to compensate \cite{hubara2016binarized}.
In contrast, fixed-point numbers \cite{shen2019q, hwang2014fixed}
and other more complex arithmetics \cite{zhao2019focused, zhanglq2018}
have larger bitwidths but offer a better task accuracy.
These quantisation methods have primarily been applied on
convolutional and fully-connected layers since they are the
most compute-intensive building blocks in CNNs and RNNs.
Another way of reducing the inference cost of DNNs
is architectural engineering.
For instance, using depth-wise separable convolutions
to replace normal convolutions not only
costs fewer parameters
but also achieves comparable accuracy \cite{howard2017mobilenets}.
However, architectural engineering is normally tedious and complex;
recent research on NAS reduces the amount of manual tuning in
the architecture design process.
Initial NAS methods used evolutionary algorithms
and reinforcement learning
to find optimal network architectures, but each iteration of the search
fully trains and evaluates
many child networks \cite{real2017large, zoph2016neural},
thus needing a huge amount of computation resources and time.
Liu \etal~proposed Differentiable Architecture Search (DARTs) that
creates a supernet with all possible operations
connected
by probabilistic priors \cite{liu2018darts}.
The search cost of DARTs is reduced by orders of magnitude
-- it is the same as training a single supernet (one-shot).
In addition, DARTs is gradient-based,
meaning that standard Stochastic Gradient Descent (SGD)
now can be used to update these probabilistic priors.
One major drawback of DARTs is that
all operations of the supernet
are active during the search.
Recently proposed single-path NAS methods 
only update a sampled network from the supernet at each search step, 
and are able to converge quickly and significantly
reduce the computation and memory cost compared to DARTs
\cite{wu2019fbnet, cai2018proxylessnas, guo2019single}.
In addition, many of the NAS methods on vision networks
consider mixed quantisation in their search space
\cite{guo2019single, wu2018mixed},
but limit the quantisation granularity to per-convolution level.

\subsection{Graph Neural Network}
Deep Learning on graphs has emerged into an important field of machine learning in recent years, partially due to the increase in the amount of graph-structured data. Graph Neural Networks has scored success in a range of different tasks on graph data, such as node classification~\cite{velickovic2018graph,kipf2016semi,hamilton2017inductive}, graph classification~\cite{zhang2018end,ying2018hierarchical,xu2018how} and link prediction~\cite{zhang2018link}.
There are many variants of graph neural networks proposed to tackle graph structured data. In this work we focus on GNNs applied to node classification tasks based on Message-Passing Neural Networks (MPNN) \cite{gilmer2017neural}.
The objective of the neural network is to learn an embedding function for node features so that they quickly generalise to other unseen nodes. 
This inductive nature is needed for large-scale production level machine learning systems, since they normally operate on a constantly changing graph with many coming unseen nodes (\eg~shopping history on Amazon, posts on Reddit \etc) \cite{hamilton2017inductive}.
It is also worth to mention that many of these systems are high-throughput and latency-critical. 
The reductions in energy and latency of the deployed networks imply the service providers can offer a better user experience while keeping a lower cost. 

Most of the manually designed GNN architectures
proposed for the node classification use MPNN. The list includes, but is not limited to,
GCN~\cite{kipf2016semi}, GAT~\cite{velickovic2018graph}, GraphSage~\cite{hamilton2017inductive}, and their variants~\cite{shi2019two,zhang2018gaan,chen2017stochastic,wang2019heterogeneous}. These works differ in ways of computing the edge weights, sampling neighbourhood and aggregating neighbour messages. We also relate to works focusing on building deeper Graph Neural Networks with the help of residual connections, such as JKNet~\cite{xu2018representation}. To the best of our knowledge, there is no prior work in investigating quantisation for Graph Neural Networks.

There is a recent surge of interest in
looking at how to extend NAS methods
from image and sequence data to graphs.
Gao \etal~proposed GraphNAS
that is a RL-based NAS applied to graph data
\cite{gao2019graphnas}.
Later, Zhou \etal~combined the RL-based NAS
with parameter sharing \cite{zhou2019auto}
and developed AutoGNN.
Zhao \etal~
proposed a gradient-based, single-shot NAS
called PDNAS,
and searched both the micro-
and macro-architectures of GNNs \cite{zhao2020probabilistic}.
All of these graph NAS methods show superior performance
when compared to other manually designed networks.

\begin{table*}[h!]
\caption{
    Quantisation search space for weights and activations.
}
\label{tab:quan_search}
\vskip 0.15in
\begin{center}
\begin{small}
\begin{sc}
\adjustbox{max width=0.75\textwidth}{%
\begin{tabular}{ccc|ccc}
\toprule
\multicolumn{3}{c}{Weights} &
\multicolumn{3}{c}{Activations} \\
\midrule
Quantisation & Frac Bits & Total Bits &
Quantisation & Frac Bits & Total Bits \\
\midrule
binary      & 0          & 1          &
fix$2.2$    & 2          & 4          \\
binary      & 0          & 1          &
fix$4.4$    & 4          & 8          \\
ternary     & 0          & 2          &
fix$2.2$    & 2          & 4          \\
ternary     & 0          & 2          &
fix$4.4$    & 4          & 8          \\
ternary     & 0          & 2          &
fix$4.8$    & 4          & 12         \\
fix$1.3$    & 3          & 4          &
fix$4.4$    & 4          & 8          \\
fix$2.2$    & 2          & 4          &
fix$4.4$    & 4          & 8          \\
fix$1.5$    & 5          & 6          &
fix$4.4$    & 4          & 8          \\
fix$3.3$    & 3          & 6          &
fix$4.4$    & 4          & 8          \\
fix$2.4$    & 4          & 6          &
fix$4.4$    & 4          & 8          \\
fix$4.4$    & 4          & 8          &
fix$4.4$    & 4          & 8          \\
fix$4.4$    & 4          & 8          &
fix$4.8$    & 8          & 12          \\
fix$4.4$    & 4          & 8          &
fix$8.8$    & 8          & 16          \\
fix$4.8$    & 8          & 12          &
fix$4.8$    & 8          & 12          \\
fix$4.12$   & 12          & 16          &
fix$4.4$    & 4          & 8          \\
fix$4.12$   & 12          & 16          &
fix$4.8$    & 8          & 12          \\
fix$4.12$   & 12          & 16          &
fix$8.8$    & 8          & 16          \\
\bottomrule
\end{tabular}
}
\end{sc}
\end{small}
\end{center}
\end{table*}

\section{Method}
\begin{figure*}[!h]
	\centering
	\includegraphics[width=.85\linewidth]{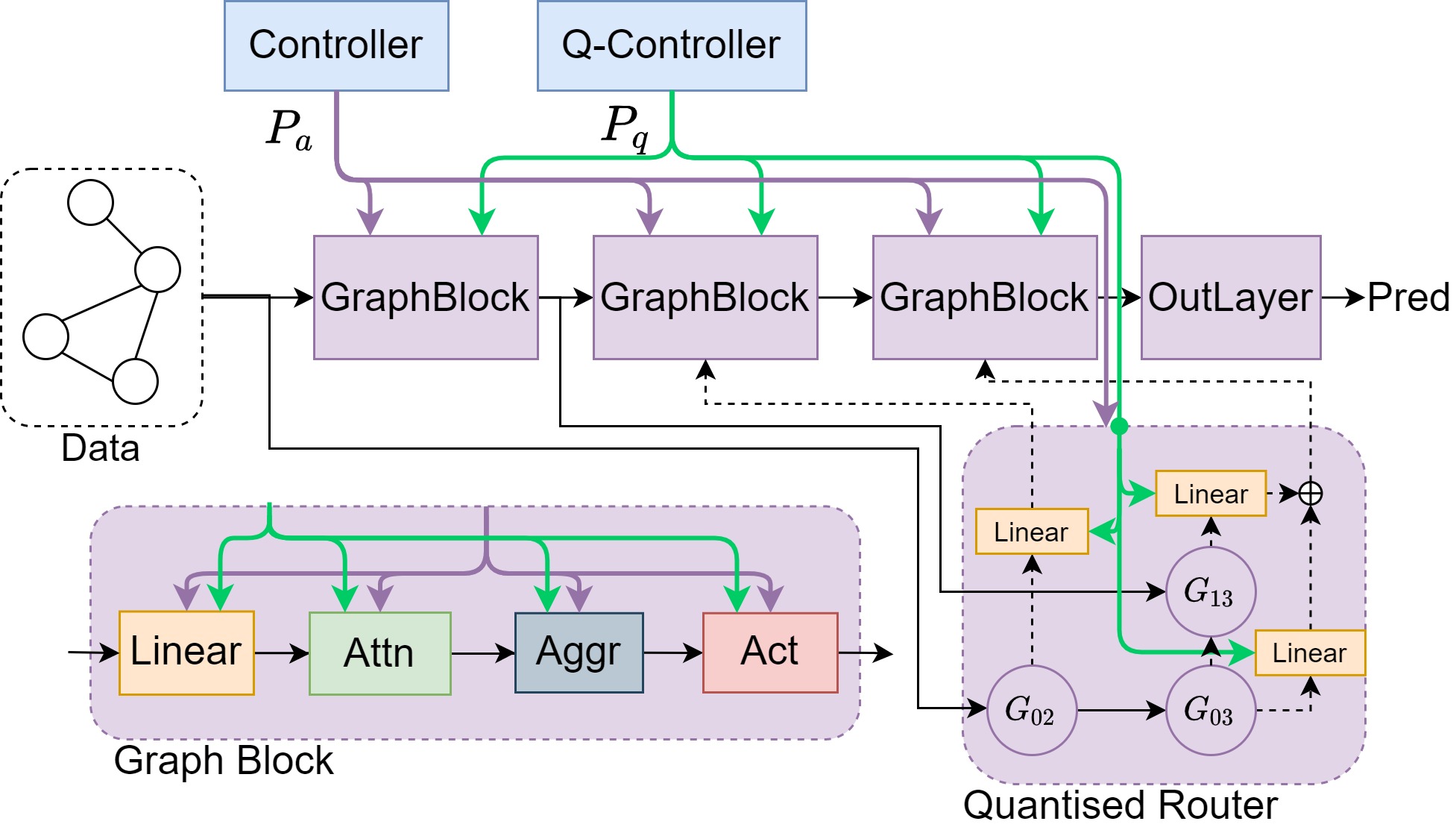}
	\caption{An overview of the LPGNAS architecture. $P_a$ denotes controller output (Purple) for selecting different operations within each graph block, and for routing shortcut connections between graph blocks. $P_q$ denotes quantisation controller (Q-Controller) output (Green) for selecting quantisations for operations within graph blocks and shortcut connections. $G_{ij}$ are gating functions conditioned on $P_a$. Solid lines are input streams into the router while dashed lines are output streams.}
	\label{fig:overview}
\end{figure*}
\subsection{Quantisation Search Space}

\begin{table}[!h]
\caption{Different types of attention mechanisms. $W$ here is parameter vector for attention. $<,>$ is dot product, $a_{ij}$ is attention for message from node $j$ to node $i$.}
\label{tab:attn_types}
\vskip 0.15in
\begin{center}
\begin{small}
\begin{sc}
\begin{tabular}{c|l}
\toprule
Attention Type & Equation\\
\midrule
Const & $a_{ij} = 1$ \\
GCN & $a_{ij} = \frac{1}{\sqrt{d_i d_j}}$\\
GAT & $a^{gat}_{ij} = \mathsf{LeakyReLU}(W_a (h_i||h_j))$ \\
Sym-GAT & $a_{ij} = a^{gat}_{ij} + a^{gat}_{ji}$\\
COS & $a_{ij} = <W_{a1} h_i, W_{a2} h_j> $\\
Linear & $a_{ij} = \mathsf{tanh}(\sum_{j \in N(i)}(W_a h_j)) $\\
Gene-Linear & $a_{ij} = W_g \mathsf{tanh}(W_{a1} h_i + W_{a2} h_j) $ \\
\bottomrule
\end{tabular}
\end{sc}
\end{small}
\end{center}
\vskip -0.1in
\end{table}

\begin{table}[!h]
\caption{Different types of activation functions.}
\label{tab:act_types}
\vskip 0.15in
\begin{center}
\begin{small}
\begin{sc}
\begin{tabular}{c|l}
\toprule
Activation & Equation\\
\midrule
None & $f(x) = x$ \\
Sigmoid & $f(x) = \frac{1}{1+e^{-x}}$\\
Tanh & $ f(x) = tanh(x)$ \\
Softplus & $ f(x) = \frac{1}{\beta} \log (1+e^{\beta x})$\\
ReLU & $f(x) = Max(0,x)$ \\
LeakyReLU & $f(x) = Max(0,x) + \alpha Min(0,x)$ \\
ReLU6 & $f(x) = Min(Max(0,x),6)$ \\
ELU & $f(x) = Max(0,x) + Min(0, \alpha (e^x-1))$ \\
\bottomrule
\end{tabular}
\end{sc}
\end{small}
\end{center}
\vskip -0.1in
\end{table}

We consider a single graph block,
or a GNN layer,
as four consecutive sub-blocks
(\Cref{equ:graphblock}).
\Cref{equ:graphblock} considers
node features $h^{k-1}$ from the $k-1$
layer as inputs,
and produces new node features $h^{k}$
with trainable attention parameters $a^{k}$
and weights $w^{k}$.
The four sub-blocks, as illustrated in \Cref{equ:graphblock}, 
are the Linear block,
Attention block, Aggregation block
and Activation block; these
sub-blocks have operations as search
candidates,
which is a similar architectural search space to
Zhao \etal~\cite{zhao2020probabilistic}
and Gao \etal~\cite{gao2019graphnas}.
We provide a detailed description of
the architectural search space
in the Appendix.
In \Cref{equ:qgraphblock},
we label all possible quantisation
candidates using \textcolor{blue}{$Q$}.
The quantisation function
\textcolor{blue}{$Q$}
can be applied not only on learnable
parameters (\eg,~$w^{k}$) but also
activation values between consecutive
sub-blocks.
In addition, we allow quantisation options
in \Cref{equ:qgraphblock}
to be different.
For instance, $\textcolor{blue}{Q_{a}}$ can learn
to have a different quantisation from $\textcolor{blue}{Q_{w}}$,
meaning that a single
graph block receives a mixed quantisation
for different quantisable
components annotated in \Cref{equ:qgraphblock}.

\begin{equation}
		h^k = \mathsf{Act}(
			\mathsf{Aggr}(\mathsf{Atten}(a^k, \mathsf{Linear}(w^{k}, h^{k-1})))) 
	\label{equ:graphblock}
\end{equation}
\begin{equation}
	\begin{split}
		h^k_{\mathsf{linear}} &= \textcolor{blue}{Q_{l}(}	
		          \mathsf{Linear}(
					\textcolor{blue}{Q_{w}(}
					    w^{k}
					\textcolor{blue}{)}
					,
					\textcolor{blue}{Q_{h}(}
					    h^{k-1}
					\textcolor{blue}{)}
					)
				\textcolor{blue}{)} \\
		h^k_{\mathsf{atten}} &= 	\textcolor{blue}{Q_{at}(}
		            \mathsf{Atten}(
					    \textcolor{blue}{Q_{a}(}
					    a^k \textcolor{blue}{)}, 
					    h^k_{\mathsf{linear}}
					)
				\textcolor{blue}{)}	\\		
		h^k &= 	\mathsf{Act}(
		        \textcolor{blue}{Q_{ag}(}
				    \mathsf{Aggr}(h^k_{\mathsf{atten}})
				\textcolor{blue}{)}
                ) \\
	\end{split}
	\label{equ:qgraphblock}
\end{equation}

The reasons for considering input activation values
as quantisation candidates are the following.
First, GNNs always consider large input graph data,
the computation operates on a per-node or per-edge resolution
but requires the entire graph or the entire sampled graph as inputs.
The amount of intermediate data produced during the computation
is huge and
causes a large pressure on the amount of on-chip memory available.
Second, quantised input activations with quantised parameters
together simplify the arithmetic operations.
For instance,
$\mathsf{Linear}( \textcolor{blue}{Q_w(} w^{k} \textcolor{blue}{)}, \textcolor{blue}{Q_h(} h^{k-1} \textcolor{blue}{)}) \textcolor{blue}{)})$
considers both quantised $h^{k-1}$ and quantised $w^k$
so that the matrix multiplication with these values can operate
on a simpler fixed-point arithmetic.

The quantisation search space identified
in \Cref{equ:qgraphblock} is
different from the search space identified by
Wu \etal~\cite{wu2018mixed} and Guo \etal~\cite{guo2019single}.
Most existing NAS methods focusing on quantising CNNs
look at quantisation at each convolutional block.
In our case, a single graph block is equivalent
to a convolutional block and we look at more fine-grained
quantisation opportunities within the four sub-blocks.
The quantisation search considers a wide range of fixed-point
quantisations.
The weights and activation can pick the numbers of bits in $[1,2,4,6,8,12,16]$ and $[4, 8, 12, 16]$ respectively, and we allow different integer and fractional bits combinations
We list the detailed quantisation options in the \Cref{tab:quan_search}.
\Cref{tab:quan_search} shows the quantisation search space,
each quantisable operation identified will have
these quantisation choices available.
BINARY means binary quantisation, and
TERNARY means two-bit tenary quantiation.
FIX$x.y$ means fixed-point quantisation
with $x$ integer bits and $y$ bits for fractions.

In total, as listed in \Cref{tab:quan_search}, we have $17$ quantisation options; and as illustrated in \Cref{equ:qgraphblock}, there are four sub-blocks that join the quantisation search, this gives us in total $17^4 = 83251$ quantisation options.

\subsection{Architecture Search Space}

As illustrated in \Cref{fig:overview}, each graph block consists of four sub-blocks, namely the linear block, attention block, aggregation block and activation block.
Each of these sub-blocks contain architectural choices that joins the NAS process.
In \Cref{tab:attn_types} and \Cref{tab:act_types}, we list all attention types and activation types that were considered
in the architectural search space.
We also consider searching for the best hidden feature
size, using two fully-connected layers with an intermediate layer with an expansion factor that can be picked from a set $\{1, 2, 4, 8\}$.
The possible aggregation choices include $mean$, $add$ and $max$.

The considered search space is similar to prior works in this domain \cite{zhao2020probabilistic, zhou2019auto, gao2019graphnas}.
The possible number of architectural combinations of a single layer is 
$7 \times 8 \times 3 \times 4 = 672$,
for an $n$-layer network, this means there are in total $672^n$ possible architectural combinations.
Combining the quantisation search space mentioned in the previous section, the search space has in total $83251 \times 672^n$ options.
Assuming the number of layers considered in the search is $n=4$, this roughly gives us a search space of $10^{15}$ options.

\subsection{Low Precision Graph Network Architecture Search}

\begin{algorithm}[!h]
	\caption{LPGNAS algorithm}
	\label{alg:lpgnas}
	\begin{algorithmic}
		\STATE {\bfseries Input:} $x$, $y$, $x_v$, $y_v$, $M$, $M_a$, $M_q$, $K$, $\alpha$, $\beta$
		\STATE $\mathsf{Init}(w, w_a, w_q)$

		\FOR{$i=0$ {\bfseries to} $M - 1$}
		    \STATE $N$ = NoiseGen($i$, $\alpha$)
		    \STATE $P_a, P_q$ = $g_a(w_a, x_{\mathsf{val}}, N), g_q(w_q, x_{\mathsf{val}}, N)$
		    \STATE $\pi_a, \pi_q = \argmax(P_a), \argmax(P_q)$
		    \FOR{$i=0$ {\bfseries to} $K - 1$}
               \STATE $\mathcal{L} = \mathsf{Loss}(x, y, \pi_a, \pi_q)$
               \STATE $w$ = $\mathsf{Opt}_w (\mathcal{L})$
		    \ENDFOR
            \STATE $\mathcal{L}_{v} = \mathsf{Loss}(x_v, y_v, \pi_a, \pi_q)$
            \IF{$e>M_a$}
                \STATE $w_a$ = $\mathsf{Opt}_{w_a} (\mathcal{L}_{v})$
            \ENDIF
            \IF{$e>M_q$}
				\STATE $\mathcal{L}_{q} = \mathsf{QLoss}(P_a, P_q)$
                \STATE $w_q$ = $\mathsf{Opt}_{w_q} (\mathcal{L}_{v} + \beta L_q)$
            \ENDIF
		\ENDFOR
	\end{algorithmic}
\end{algorithm}

We describe the Low Precision Graph Network Architecture Search (LPGNAS)
algorithm in Algorithm \ref{alg:lpgnas}.
($x$, $y$), ($x_v$, $x_v$)
are the training and validation data respectively.
$M$ is the total number of search epochs,
and $M_a$ and $M_q$ are the epochs to start
architectural and quantisation search.
Similar to Zhao \etal~\cite{zhao2020probabilistic}, before the number of epochs reaches $M_a$ or $M_q$,
LPGNAS randomly picks up samples and warms up the trainable parameters in the supernet.
$K$ is the number of steps to iterate in training the supernet.
After generating noise $N$,
we use this noise in architectural controller $g_a$
and quantisation controller $g_q$ together with the validation data $x_v$ to determine the architectural $\pi_a$ and quantisation $\pi_q$ choices. After reaching the pre-defined starting epoch ($M_a$ or $M_q$), LPGNAS starts to optimise the controllers' weights ($w_a$ and $w_q$).
We choose the hyper-parameters
$M_q=20, M_a=50, \alpha=1.0, \beta=0.1$,
unless specified otherwise,
based on the
choices made
by Zhao \etal~\cite{zhao2020probabilistic}.
In addition, we provide an ablation study in the Appendix
to justify our choices of hyper-parameters.
Notice that $\mathsf{QLoss}$
estimates the current hardware cost based on both
architectural and quantisation probabilities.
As shown in \Cref{equ:qloss},
we define $S$ as a set of values that represents
the costs (number of parameters) of all possible architectural options:
for each value $s$ in this set, we multiply it with the probability $p_a$
from architecture probabilities $P_a$ generated by the NAS architecture controller.
Similarly, we produce the weighted-sum for quantisation from
the set of values $S_q$ that represents costs of all possible quantisations.
The dot product $\cdot$
between these two weighted-sum vectors produces the quantisation loss $L_q$.
\begin{equation}
	\begin{split}
		L_q & = \mathsf{QLoss}(P_a, P_q) \\
			& = \sum_{s \in S, p_a \in P_a}(s * p_a) \cdot \sum_{s_q \in S_q, p_q \in P_q}(s_q * p_q)
	\end{split}
	\label{equ:qloss}
\end{equation}

\Cref{fig:overview} is an illustration of the LPGNAS algorithm. We use two controllers ($g_a$ and $g_q$)
for architecture and quantisation
(named as Controller and Q-Controller in the diagram) respectively.
The controllers use trainable
embeddings connected to linear layers
to produce architecture and quantisation probabilities.
The architecture controller
provides probabilities ($P_a$) for picking
architectural options of graph blocks and also the router.
The router is in charge of determining how
each graph block shortcuts to the subsequent graph blocks \cite{zhao2020probabilistic}.
In addition, the quantisation controller
provides quantisation probabilities ($P_q$) to both modules
in graph blocks and the linear layers in the router.
In a graph block, only the linear and attention
layers have quantisable parameters and have
matrix multiplication operations;
neither activation nor aggregation
layers
have any trainable or quantisable parameters.
However, all input activation values of each layer
in the graph block are quantised.
The amount of intermediate data produced
during the computation causes memory pressure
on many existing or emerging hardware devices,
so even for modules that do not have quantisable parameters,
we choose to quantise their input activation values
to help reduce the amount of buffer space required.


\section{Results}

\begin{table*}[t!]
\caption{
    Accuracy and size comparison on
    Cora, Pubmed and Citeseer with
    data splits same as \citet{yang2016revisiting}.
    Our results are averaged across 3 independent runs.
    The numbers in bold show best accuracies or smallest sizes.
}
\label{tab:citation:original}
\vskip 0.15in
\begin{center}
\begin{small}
\begin{sc}
\adjustbox{max width=\textwidth}{%
\begin{tabular}{cc|cccccc}
\toprule
\multirow{2}*{Method}   & \multirow{2}*{Quan}    & \multicolumn{2}{c}{Cora} & \multicolumn{2}{c}{CiteSeer} & \multicolumn{2}{c}{PubMed} \\
                        &        & Accuracy           & Size         & Accuracy           & Size        & Accuracy   & Size  \\
\midrule
GraphSage               & float  & $74.5 \pm 0.0 \%$  & $92.3$KB     & $75.3 \pm 0.0 \%$  & $237.5$KB   & $85.3 \pm 0.1 \%$  & $32.2$KB \\
GraphSage               & w10a12 & $74.3 \pm 0.1 \%$  & $28.8$KB     & $75.1 \pm 0.1 \%$  & $74.2$KB    & $85.0 \pm 0.0 \%$  & \cellcolor{red!25} $10.1$KB  \\
GAT                     & float  & $88.9 \pm 0.0 \%$  & $369.5$KB    & $75.9 \pm 0.0 \%$  & $950.3$KB   & $86.1 \pm 0.0 \%$  & $129.6$KB \\
GAT                     & w4a8   & $88.8 \pm 0.1 \%$  & $46.2$KB     & $68.0 \pm 0.1 \%$  & $118.8$KB   & $82.0 \pm 0.0 \%$  & $16.2$KB  \\
JKNet                   & float  & $88.7 \pm 0.0 \%$  & $214.9$KB    & $75.5 \pm 0.0 \%$  & $505.2$KB   & $87.6 \pm 0.0 \%$  & $94.5$KB  \\
JKNet                   & w6a8   & $88.7 \pm 0.1 \%$  & \cellcolor{red!25}$\mathbf{40.3}$KB     & $73.2 \pm 0.1 \%$  & $94.7$KB    & $86.1 \pm 0.1 \%$  & $17.7$KB  \\
\midrule
PDNAS-2                 & float  & $89.3 \pm 0.1 \%$ & $192.2$KB     & \cellcolor{blue!25} $\mathbf{76.3 \pm 0.3 \%}$ & $478.6$KB    & $ 89.1 \pm 0.2 \%$  & $72.8$KB  \\
PDNAS-3                 & float  & $89.3 \pm 0.1 \%$ & $200.0$KB     & $75.5 \pm 0.3 \%$ & $494.4$KB    & $ 89.1 \pm 0.2 \%$  & $81.4$KB   \\
PDNAS-4                 & float  & $89.8 \pm 0.3 \%$ & $205.0$KB     & $75.6 \pm 0.2 \%$ & $500.0$KB    & $ 89.2 \pm 0.1 \%$  & $102.7$KB  \\
\midrule
LPGNAS                  & mixed  & \cellcolor{blue!25} $\mathbf{89.8 \pm 0.0 \%}$ & $67.3$KB      & $76.2 \pm 0.1\%$  &  \cellcolor{red!25} $\mathbf{56.5}$KB    & \cellcolor{blue!25}$ \mathbf{89.6 \pm 0.1 \%}$  & $45.6$KB  \\
\bottomrule
\end{tabular}
}
\end{sc}
\end{small}
\end{center}
\end{table*}

\begin{figure*}[!ht]
	\begin{center}
        \includegraphics[width=\linewidth]{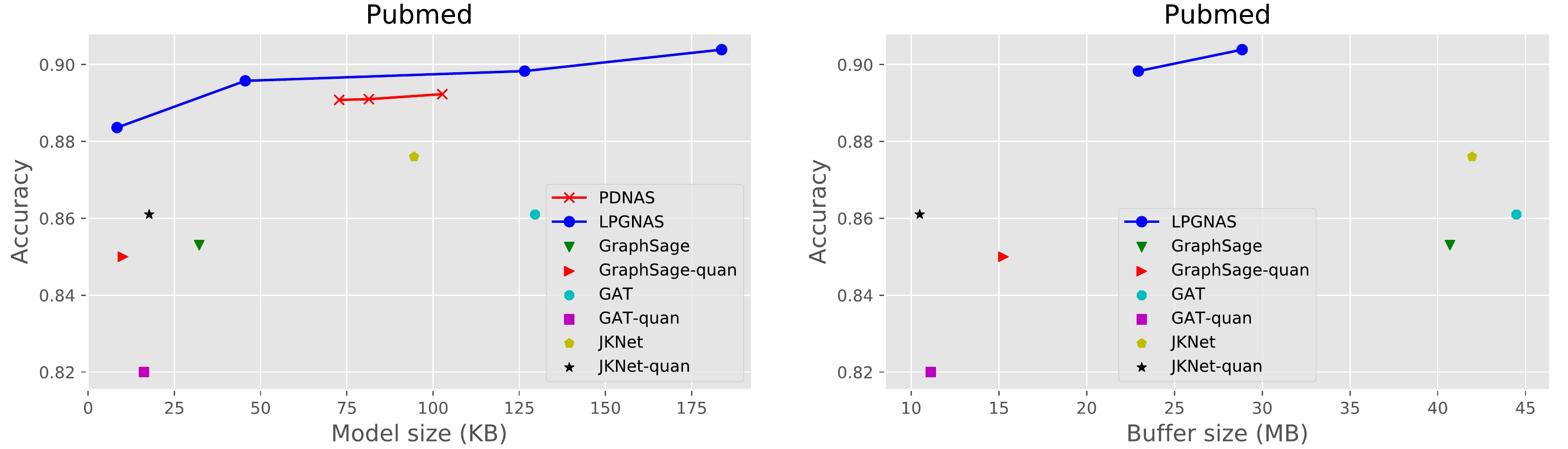}
	\end{center}
	\caption{
		Pareto Frontier of LPGNAS and PDNAS \cite{zhao2020probabilistic}
		on the Pubmed datset \cite{yang2016revisiting},
		the manually designed networks are shown as dots.
		On the left, we show the trade-off between model sizes and accuracy; on the right, the trade-off is between buffer sizes and accuracy.
		}
	\label{fig:pubmed}
\end{figure*}

We compare LPGNAS to both manually designed and
searched GNNs.
In addition, we provide comparisons
for two types of tasks.
The first type is on the Citation datasets,
where the network takes the entire graph as an input
in a transductive learning setup \cite{yang2016revisiting}.
The second type considers inputs sampled from a large graph,
and we use
the inductive GraphSAINT sampling strategy \cite{zeng2020graphsaint}.
For the networks to learn from sampled graphs,
we present results on larger datasets, including
Amazon Computers,
Amazon Photos \cite{shchur2018pitfalls},
Yelp, Flickr \cite{zeng2020graphsaint}
and Cora-full \cite{bojchevski2017deep}.

\begin{table*}[h!]
\caption{
    Accuracy and size comparison on
    Amazon-Computers and Amazon-Photo \cite{shchur2018pitfalls}
    Our results are averaged across 3 independent runs.
    The numbers in bold show best accuracies (blue) and smallest model sizes (red).
}
\label{tab:sampled}
\vskip 0.15in
\begin{center}
\begin{small}
\begin{sc}
\adjustbox{max width=0.7\textwidth}{%
\begin{tabular}{cc|cccc}
\toprule
\multirow{2}*{Method}
& \multirow{2}*{Quan}
& \multicolumn{2}{c}{Amazon-Computers}
& \multicolumn{2}{c}{Amazon-Photos} \\
&
& Accuracy     & Size
& Accuracy     & Size \\
\midrule
GAT     & float &
$84.4 \pm 0.3\%$    &199.8KB    &
$84.2 \pm 0.1\%$    &199.8KB    \\
GAT    & w4a8 &
$81.8 \pm 1.5\%$    &25.0KB     &
$83.5 \pm 0.5\%$    &25.0KB     \\
GAT-V2   & float &
$85.3 \pm 0.3\%$    &1597.6KB   &
$85.2 \pm 0.4\%$    &1597.6KB   \\
GAT-V2  & w4a8
&$38.0 \pm 0.0\%$   &199.7KB
&$38.0 \pm 0.0\%$   &199.7KB \\
\midrule
JKNet   & float
&$87.6 \pm 0.0\%$   &130.5KB
&$87.7 \pm 0.1\%$   &130.5KB \\
JKNet  & w4a8
&$38.0 \pm 0.0\%$   &16.3KB
&$38.0 \pm 0.0\%$   &16.3KB \\
JKNet-V2& float
&$88.8 \pm 0.1\%$    &8968.2KB
&$88.6 \pm 0.1\%$    &8968.2KB \\
QJKNet-V2& w4a8
&$38.0 \pm 0.0\%$   &1121.0KB
&$38.0 \pm 0.0\%$   &1121.0KB \\
\midrule
SageNet& float
&$84.0 \pm 0.0\%$   &49.8KB
&$84.0 \pm 0.1\%$   &49.8KB \\
SageNet& w4a8
&$83.9 \pm 0.1\%$ &6.2KB
&$83.9 \pm 0.1\%$ &6.2KB \\
SageNet-V2& float
&$87.7 \pm 0.2\%$   &1593.4KB
&$87.9 \pm 0.2\%$   &1593.4KB \\
SageNet-V2& w4a8
&$87.3 \pm 0.2\%$ &199.2KB
&$87.2 \pm 0.3\%$ &199.2KB \\
\midrule
LPGNAS & mixed
&\cellcolor{blue!25}$ \mathbf{90.5 \pm 0.0 }\%$   & 157.3KB
&\cellcolor{blue!25}$ \mathbf{89.7 \pm 0.0 }\%$   & 164.0KB\\
LPGNAS-small & mixed
&$  88.7 \pm 0.1 \%$  &\cellcolor{red!25} $\mathbf{3.5}$KB
&$  88.6 \pm 0.1 \%$  &\cellcolor{red!25} $\mathbf{3.6}$KB \\
\bottomrule
\end{tabular}
}
\end{sc}
\end{small}
\end{center}
\end{table*}

\begin{table*}[h!]
\caption{
    Accuracy and size comparison on
    Flickr \cite{zeng2020graphsaint},
    Cora-full \cite{bojchevski2017deep} and Yelp \cite{zeng2020graphsaint}.
    Our results are averaged across 3 independent runs.
    The numbers in bold show best accuracies (blue) and smallest model sizes (red).
}
\label{tab:sampled2}
\vskip 0.15in
\begin{center}
\begin{small}
\begin{sc}
\adjustbox{max width=1\textwidth}{%
\begin{tabular}{cc|cccccc}
\toprule
\multirow{2}*{Method}
& \multirow{2}*{Quan}
& \multicolumn{2}{c}{Flickr}
& \multicolumn{2}{c}{CoraFull}
& \multicolumn{2}{c}{Yelp} \\
&
& Accuracy     & Size
& Accuracy     & Size
& Micro-F1           & Size \\
\midrule
GAT     & float &
$42.4 \pm 0.1\%$    &130.6KB    &
$62.0 \pm 0.1\%$    &2249.3KB   &
$24.6 \pm 0.5\%$    &104.4KB    \\
QGAT    & w4a8 &
$42.4 \pm 0.0\%$    &16.3KB     &
$53.8 \pm 0.2\%$    &281.2KB    &
$14.2 \pm 0.0\%$    &13.0KB     \\
GAT-V2   & float &
$44.6 \pm 0.9\%$    &1044.6KB   &
$65.5 \pm 0.3\%$    &17988.4KB  &
$32.0 \pm 0.0\%$    &826.5KB    \\
QGAT-V2  & w4a8  &
$42.3 \pm 0.0\%$    &130.6KB
&$4.4 \pm 0.0\%$    &2248.6KB
&$14.2 \pm 0.0\%$   &103.3KB \\
\midrule
JKNet   & float  &
$50.6 \pm 0.0\%$    &95.5KB
&$69.0 \pm 0.1\%$   &1162.8KB
&$32.6 \pm 0.3\%$   &94.1KB \\
QJKNet   & w4a8
&$42.3 \pm 0.0\%$   &11.9KB
&$4.4 \pm 0.0\%$    &145.3KB
&$31.0 \pm 2.2\%$   &11.8KB \\
JKNet-V2& float
&$50.2 \pm 0.2\%$    &8409.1KB
&$68.7 \pm 0.1\%$    &25481.5KB
&$33.1 \pm 1.6\%$    &8380.8KB \\
QJKNet-V2& w4a8
&$42.3 \pm 0.0\%$    &1051.1KB
&$4.4 \pm 0.0\%$     &3185.2KB
&$14.2 \pm 0.0\%$    &1047.6KB \\
\midrule
SageNet    &  float
& $49.9 \pm 0.0\%$   &32.5KB
& $66.4 \pm 0.1\%$   &562.3KB
& $23.6 \pm 0.0\%$   &26.1KB \\
QSageNet   & float
&$45.0 \pm 0.0\%$    &4.1KB
&$65.7 \pm 0.2\%$    &70.3KB
&$24.8 \pm 0.0\%$    &3.3KB \\
SageNet-V2 &  float
&$50.0 \pm 0.0\%$    &1040.4KB
&$68.2 \pm 0.0\%$    &17983.8KB
&$30.5 \pm 0.6\%$    &821.6KB \\
QSageNet-V2 & w4a8
&$49.7 \pm 0.3\%$    &130.1KB
&$68.4 \pm 0.3\%$    &2248.0KB
&$27.4 \pm 0.7\%$    &102.7KB \\
\midrule
LPGNAS & mixed
&\cellcolor{blue!25} $\mathbf{73.4 \pm 0.0}\%$    &81.2KB
&\cellcolor{blue!25} $\mathbf{85.5 \pm 0.0}\%$    &432.3KB
&\cellcolor{blue!25} $\mathbf{40.5 \pm 0.1}\%$    & 6.4KB\\
LPGNAS-small & mixed
&$ 70.6 \pm 0.1\%$    &\cellcolor{red!25} $\mathbf{2.2}$KB
&$ 80.2 \pm 0.1\%$    &\cellcolor{red!25} $\mathbf{70.2}$KB
&$ 30.9 \pm 0.1\%$    &\cellcolor{red!25} $\mathbf{1.5}$KB \\
\bottomrule
\end{tabular}
}
\end{sc}
\end{small}
\end{center}
\end{table*}

\begin{figure*}[!h]
	\begin{center}
        \includegraphics[width=\linewidth]{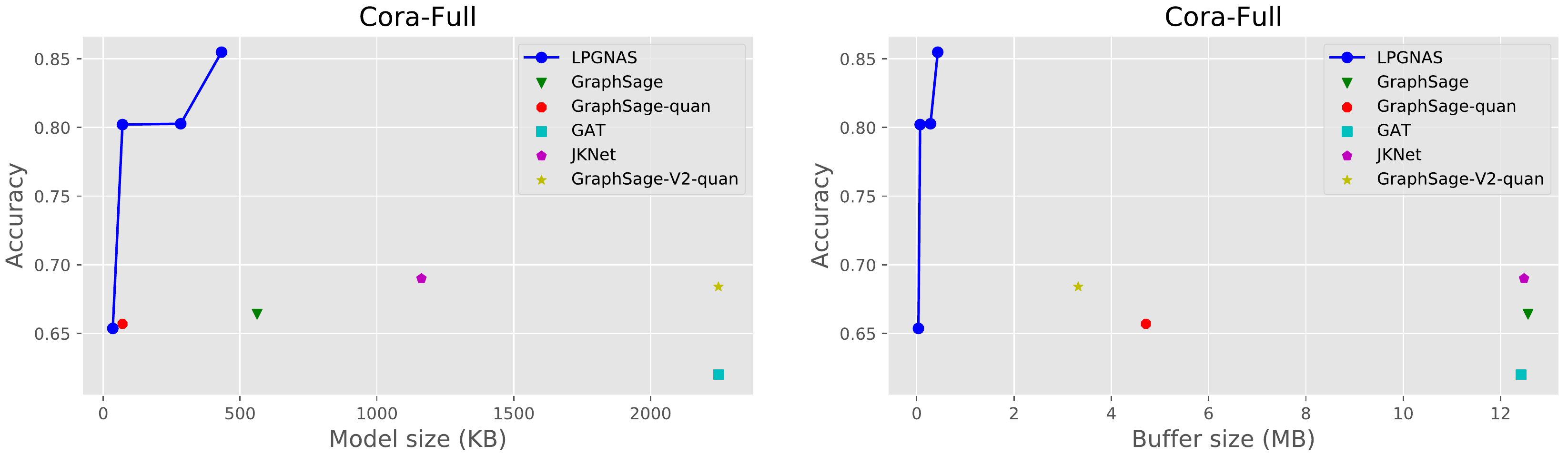}
	\end{center}
	\caption{
		Pareto Frontier of LPGNAS and other baselines
		on the Cora-full dataset \cite{yang2016revisiting},
		the manually designed networks are shown as dots.
		On the left, we show the trade-off between model sizes and accuracy; on the right, the trade-off is between buffer sizes and accuracy.
		}
	\label{fig:cora}
\end{figure*}

The Citation datasets include
nodes representing documents and edges representing citation links.
The task is to distinguish which research field the document belongs to \cite{yang2016revisiting}.
The Flickr dataset includes nodes as images, and edges exist between two images if they
share some common properties. The task is to classify these images
based on their bag-of-words encoded features \cite{zeng2020graphsaint}.
Yelp considers users as nodes and edges are made based on user friendships,
and features are word embeddings extracted from user reviews.
The task is to distinguish the categories of the shops the users
are likely to review; since multiple categories are possible, the labels are multi-hot \cite{zeng2020graphsaint}.
Amazon Computers and Photo
are part of the Amazon co-purchase graph \cite{mcauley2015image}.
The nodes are goods, and edges are the frequency between two goods bought together.
Node features are bag-of-words encoded product reviews,
and classes are the product categories.
Cora-full is an extended version of Cora, which one of the dataset in
the Citation datasets.
We provide a detailed overview of the properties and statistics of the datasets in Appendix.

\subsection{Citation Datasets}
\Cref{tab:citation:original} shows
the performance of
GraphSage \cite{hamilton2017inductive},
GAT \cite{velickovic2018graph},
JKNet \cite{xu2018representation},
PDNAS \cite{zhao2020probabilistic}
and LPGNAS on the
citation datasets
with a partition of $0.6$, $0.2$, $0.2$
for training, validation and testing respectively,
which is
the same training setup as in Yang \etal~\cite{yang2016revisiting}.
For the quantisation options of
GraphSage, GAT and JKNet,
we manually perform a grid search on the Cora dataset.
The grid search considers various weight
and activation quantisations
in a decreasing order.
The search
terminates when quantisation
causes a decrease in accuracy bigger than $0.5\%$
and rolls back to pick the previous quantisation applied as the result.
We reimplemented
GraphSage \cite{hamilton2017inductive},
GAT \cite{velickovic2018graph} and
JKNet \cite{xu2018representation} for quantisation,
and provide the architecture choices and
details of the grid search in the Appendix.
For the manual baseline networks,
we used the quantisation strategy found on
Cora for Citeseer and Pubmed.
For LPGNAS, we searched
with a set of base configurations
that we clarified in the Appendix.
For all reimplemented baselines,
we train the networks with three different
seeds and report
the averaged results
with standard deviations.
For LPGNAS, we run the entire
search and train cycle three times and report the
final averaged accuracy trained on the searched network.

The results in \Cref{tab:citation:original}
suggest that LPGNAS shows better accuracy
on both Cora and Pubmed with quantised networks.
In addition, although LPGNAS does not show the smallest
sizes on these two datasets, it is only slightly bigger
than the manual baselines but shows a much better accuracy.
On Citeseer, LPGNAS only shows slightly worse accuracy
(around $0.1\%$ less)
with a considerably smaller size (around $9\times$ reduction in model sizes).

To further prove the point that LPGNAS generates
more efficient networks, we sweep across
different configurations and different quantisation
regularisations to produce \Cref{fig:pubmed}
to visulise the performance of LPGNAS
and how it performs with small model sizes.
We explain in the Appendix how this sweep of different LPGNAS configurations work, the sweep mainly considers a combination of different configurations of how many layers and how many base channel counts are there for the backbone network to use for LPGNAS.
We plot the Pareto frontiers of LPGNAS putting
model sizes on the horizontal axis and accuracy
on the vertical axis
for Pubmed (\Cref{fig:pubmed}).
In this plot, we also demonstrate how buffer size
trades-off with accuracy.
For buffer size, it means the
total size required to hold
all intermediate activation values
during the computation of a single inference run with a batch size of one.
GNNs normally use large graphs as input data;
the amount of memory required to
hold all intermediate values
might be a limiting factor
in batched inference.
Since LPGNAS uses quantisation not only on weights
but also on input activation values, it shows
a better trade-off than the
floating-point baselines.
For \Cref{fig:pubmed},
models staying at the top left of the plots are
considered as better, since they are consuming less
hardware resources and maintaining high accuracy.
In addition, in \Cref{fig:pubmed},
we compare PDNAS \cite{zhao2020probabilistic} to
LPGNAS and found that we outperformed
the floating-point NAS methods by staying at the top left in the plot.

\begin{table*}[h]
\centering
\caption{
 Search cost in GPU hours, all experiments are conducted on an NVIDIA GeForce RTX 2080 Ti GPU.}
\begin{adjustbox}{scale=1.0,center}
\begin{tabular}{l|cccccccc}
\toprule
Dataset     & Cora  & Citeseer  & Pubmed    & Cora-full  & Flickr    & Yelp  & Amazon-photos   & Amazon-computers \\
\midrule
LPGNAS      & 3.2   & 3.6       & 4.2       & 11.8       & 11.0      & 11.8  & 11.4            & 11.2 \\
JKNet-32    & 6518.5 & 8165.6 & 5289.1      & 2180.6    & 3062.1    & 9116.7 & 1739.8      & 1786.2 \\
\bottomrule
\end{tabular}
\end{adjustbox}

\label{tab:search_time}
\end{table*}

\begin{figure*}[!h]
	\begin{center}
        \includegraphics[width=.7\linewidth]{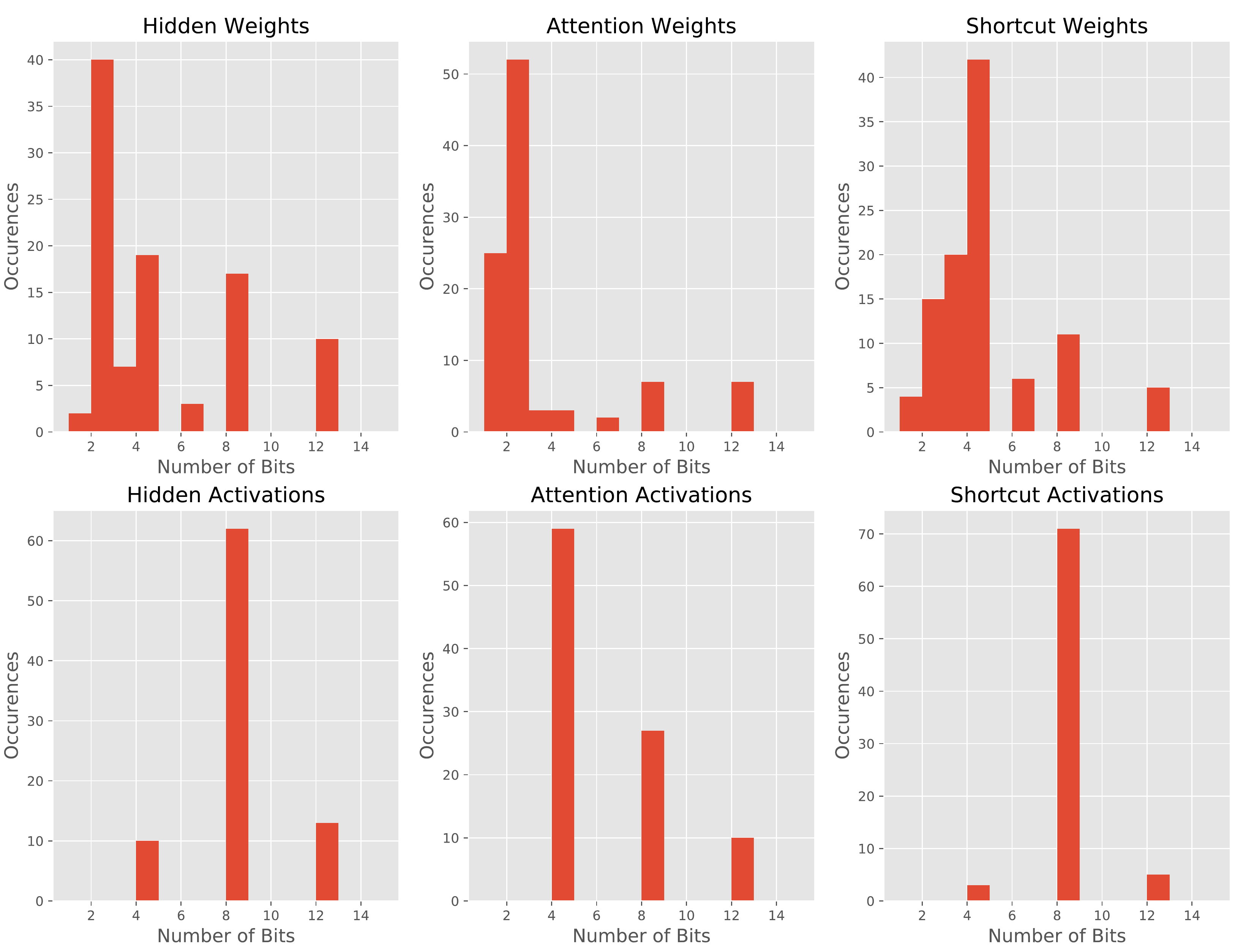}
	\end{center}
	\caption{
		Collected statistical information for quantisation,
		the horizontal axis shows the chosen bitwidth and the vertical
		axis shows the occurences. A bitwith count of 16 is not present because it has never been selected by LPGNAS.
		}
	\label{fig:quan}
\end{figure*}

\subsection{Graph Sampling Based Learning}

\Cref{tab:sampled}
and \Cref{tab:sampled2}
show how LPGNAS
performs on large datasets sampled
using the GraphSaint sampling \cite{zeng2020graphsaint}.
The detailed configuration of the sampler is in the Appendix.
We provide additional baselines on these datasets
since the original
GraphSage \cite{hamilton2017inductive},
GAT \cite{velickovic2018graph},
JKNet \cite{xu2018representation}
underfit this challenging task.
We thus implemented V2 versions of the baselines that
have larger channel counts;
the architecture details are reported in our Appendix.
In addition, for quantising the baseline models
we uniformly applied a w4a8 (4-bit weights, 8-bit input activations)
quantisation.
Although we train on a sampled sub-graph,
evaluating networks requires
a full traversal of the complete graph.
Since the datasets considered are large,
traversing the entire graph
adds a huge computation overhead.
If we use the previous
grid-search for all quantisation possibilities
(roughly $16^4$ in our case) for the all baseline networks,
this will require a huge amount of search time.
So we fixed our quantisation strategy to w4a8 for the baseline networks.

The results in \Cref{tab:sampled}
and \Cref{tab:sampled2}
suggest that LPGNAS found the models with
best accuracy and also with a relatively competitive
model size when compared to a wide range of manual
baselines.
To further demonstrate that LPGNAS is a flexible
NAS method, we adjust the base configurations to provide
models that are equal or smaller than the smallest
baseline networks in \Cref{tab:sampled} and \Cref{tab:sampled2} and
name them LPGNAS-small.
We describe how we turn the knobs
in LPGNAS by adjusting the base configurations in our Appendix.
It is clear, in both \Cref{tab:sampled} and \Cref{tab:sampled2},
that LPGNAS outperforms all baselines in terms of
accuracy or micro-F1 scores.
LPGNAS-small, on the other hand, 
trades-off the accuracy for a better model size, however
it shows better
performance than baseline models that
have a similar size budget.

We also visualise the performance of LPGNAS
on sampled datasets using the
Pareto frontiers plot for Cora-full in \Cref{fig:cora}.
As expected, LPGNAS shows a frontier at the top left,
proving that it is able to
produce small networks with high accuracy values.
In other words, LPGNAS is Pareto dominated compared other manually designed networks.

It is worth to note that LPGNAS show great performance gains compared to the baseline networks.
On Flickr, CoraFull, and Yelp, we show on average around $20\%$ increase compared to even the full-precision baseline models, while our LPGNAS produced quantised models.
The performance gap between quantised baselines and LPGNAS is even greater.
On the other hand, LPGNAS is able to produce extremely small models.
For instance, LPGNAS-Small is able to produce models that are around $50 \times$ smaller on Amazon-Computers and Amazon-Photos while only suffer from an around $1\%$ drop in accuracy compared to standard LPGNAS.

\subsection{Quantisation Search Time}
One advantage of using Network Architecture Search is the reduction of search time.
Previous NAS methods on GNNs demonstrated the effectiveness of their methods by reducing the amount of search time required to find the best network architecture \cite{zhao2020probabilistic, gao2019graphnas, zhou2019auto}.
In this work, we not only reduced the amount of manual tunning time required for network architectures but also shortened the amount of time required to search for numerical precisions at different sub-blocks.

To further illustrate that LPGNAS has significantly reduced the amount of search time required,
we provide \Cref{tab:search_time} to show how LPGNAS compares to a fixed JKNet-32 architecture in terms of the amount of GPU hours spent for searching for the best quantisation options. 
Because of the limited resources we have, we estimate the quantisation search cost of a JKNet-32 by running 5 different randomly selected quantisation combinations and multiply the averaged training time with the total number of quantisation options. 

\Cref{tab:search_time} shows that LPGNAS can significantly reduce the search time.
For instance, on Citeseer, the search time can be reduced by around $2270\times$.
It is worth noting that, when performing this quantisation search time comparison, we do not consider the architectural search space of the baseline, meaning that the baseline (JKNet) is a fixed-architecture. LPGNAS also searched for architecture choices, which would further increases the search time of the baseline if they are considered. 

\subsection{Quantisation Statistics and Limitations}
In order to
fully understand the properties of
different quantisable sub-blocks in GNNs,
we report the searched quantisation strategies
on more than 100 search runs across 8 different datasets
(Cora, Pubmed, Citeseer,
Amazon-Computers, Amazon-Photos, Flickr, CoraFull and Yelp).
The idea is to reveal some common
quantisation properties or trends using
LPGNAS on different datasets.
We collect the quantisation decisions made by LPGNAS
for both weights and activations
and present them in \Cref{fig:quan}.
The possible quantisation levels for
weights and activations
are [1, 2, 4, 6, 8, 12, 16] and [4, 8, 12, 16] bits respectively.

For weights, most quantisation
choices tend to use small bitwidths:
binary and ternary values seem to be sufficient
for the hidden and attention sub-blocks,
however, shortcut connections
prefer a larger bitwidth (around 4 bits)
for weights.
In contrast, quantising activations show in general
a trend of using larger bitwidths (around 8 bits).
Based on these results,
if one would like to manually quantise
GNNs, we would suggest to start with around 4 bits
for weights and 8 bits for activations.
However, it is worth mentioning that
LPGNAS might trade quantisation choices with
architectural decisions.
For instance, we observed LPGNAS
was able to choose operations with less
parameters but use higher bitwidths.
Nevertheless, the large number of
runs across a relatively wide range of datasets aim at sharing empirical insights for people that are manually tuning quantisations for GNNs.
For GNN services that are whether energy, memory or latency
critical, and for future AI ASIC designers focusing on GNN inference, the gathered quantisation statistics can be a useful
guideline for fitting a suitable quantisation method to their networks.

It is also interesting to observe that in both \Cref{tab:sampled} and \Cref{tab:sampled2}, some of the manually designed GNNs suffer from a large accuracy drop when applying a 4-bit weight and 8-bit activation fixed-point quantisation (w4a8).
For instance, both quantised JKNet and JKNet-V2 drop down to $4.4\%$ accuracy on the Cora-full dataset (\Cref{tab:sampled}).
However, the collected statistics of LPGNAS suggest that w4a8 is sufficient, proving that modifications of network architectures are the key element for networks to maintain accuracy with aggressive quantisation strategies.

The applied fixed-point quantisation is a straight-forward one, many researchers have proposed more advanced quantisation schemes on CNNs or RNNs \cite{zhao2019focused, zhanglq2018,shin2016fixed}.
We suggest future research of investigating how these quantisation strategies work on GNNs should consider w4a8 fixed-point quantisation as a baseline case. Because our collected statistics suggest that w4a8 fixed-point quantisation is the most aggressive quantisation for searched GNNs while guaranteeing minimal loss of accuracy, and manual GNNs often do not perform well using the w4a8 setup.
\section{Conclusion}
In this paper, we propose a novel single-path, one-shot and gradient-based NAS algorithm named LPGNAS.
LPGNAS is able to generate compact quantized networks with state-of-the-art accuracy. 
To our knowledge, this is the first piece of work studying the quantisation of GNNs. 
We define the GNN quantisation search space and show how it can be co-optimised with the original architectural search space.
In addition, we empirically show the limitation of GNN quantisation using the proposed NAS algorithm, most of the searched networks converge to a 4-bit weight 8-bit activation quantisation setup.
Despite this limitation, LPGNAS demonstrates superior performance when compared to networks generated manually or by other NAS methods on a wide range of datasets.
\nocite{langley00}

\clearpage
\bibliography{references}
\bibliographystyle{mlsys2020}

\clearpage
\appendix

\begin{figure*}[!h]
	\begin{center}
        \includegraphics[width=\linewidth]{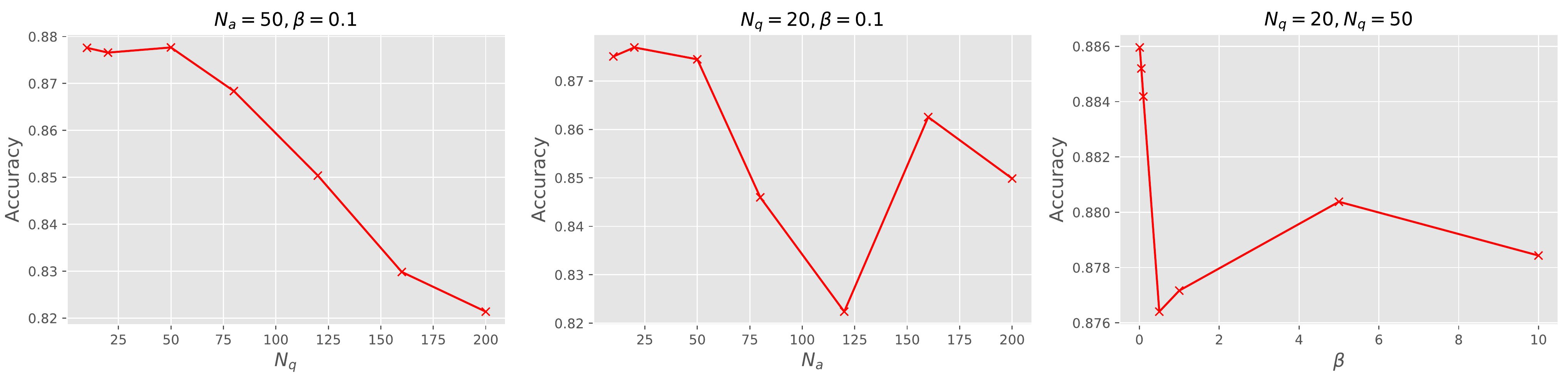}
	\end{center}
	\caption{
		Collected statistical information for quantisation,
		the horizontal axis shows the chosen bitwidth and the vertical
		axis shows the occurences.
		}
	\label{fig:hyper}
\end{figure*}
\section{Parameter Choices}
As mentioned in the LPGNAS algorithm,
we pick
$N_q=20, N_a=50, \alpha=1.0, \beta=0.1, \mathsf{lr}=0.005$.
The values of $\alpha$ and $\mathsf{lr}$
are the same
as Zhao \etal~\cite{zhao2020probabilistic}.
For $\beta$, $N_q$ and $N_a$,
we justify our choices in \Cref{fig:hyper}
by sweeping across different values of
$\beta$, $N_a$ and $N_q$
on the Pubmed dataset.

\section{Datasets Information and Data Sampler Configurations}
In this section we decribe in details the dataset we used. In our experiments we use dataset preprocessing and loading implementations from Pytorch Geometric~\cite{fey2019fast}.

\textbf{Citation Dataset}:
Citation dataset~\cite{yang2016revisiting} is a standard benchmark dataset for graph learning. It comprises three publication datasets, which are Cora, CiteSeer and PubMed. There is an additional extended version of Cora called Cora-Full~\cite{bojchevski2017deep}. Cora contains all Machine Learning papers in the Cora-Full graph. In these datasets, nodes correspond to bag-of-word features of the documents and edges indicates citation. Each node has a class label. In this paper we use the 6:2:2 train/validation/test split ratio. 

\textbf{Amazon Dataset}:
Amazon datasets~\cite{shchur2018pitfalls}, including Amazon Photos and Amazon Computers, are segments of the Amazon co-purchase graph. Nodes represent goods while edges represent frequent co-purchase of linked goods. Node feature is bag-of-word features of product review, while node label is its category. We use the 6:2:2 train/validation/test split ratio. 

\textbf{Flickr and Yelp Dataset}: Flickr and Yelp datasets are introduced along GraphSAINT~\cite{zeng2020graphsaint}. In Flickr dataset, nodes represent images uploaded to Flickr, and edges represent sharing of common properties (e.g. location and comments by same user). Node feature is 500-dimensional bag-of-words representation based on SIFT descpritions, and node label is its class. In Yelp dataset, nodes represent users and edges represent friendship between users. Node feature is summed word2vec embeddings of words in the user's reviews, and node label is a multi-hot vector representing which types of businesses has the user reviewed. For both dataset we use the 6:2:2 train/validation/test split ratio. 

\section{Grid Search and Baseline Networks Details}
For producing quantised baseline networks,
we manually grid searched options listed
in \Cref{tab:quan_search} and follow the order from bottom to top.
We stop the search and retrieve to the previous quantisation
stage if the current stage shows an accuracy drop of more than $0.5\%$.
It is worth to mention this grid search for quantisation
is very time-consuming, and we therefore only performed on Cora
and used the found quantisation strategy for the rest of the Citation datasets.

\Cref{tab:baseline} shows the configurations of the baseline
networks we've used in this paper.
\begin{table}[h!]
\caption{
    Baseline networks configurations.
}
\label{tab:baseline}
\vskip 0.15in
\begin{center}
\begin{small}
\begin{sc}
\adjustbox{max width=1\textwidth}{%
\begin{tabular}{c|cc}
\toprule
Networks &
Layers &
Channels  \\
\midrule
GAT         & 2     & 32\\
GAT-V2      & 2     & 64\\
JKNet       & 2     & 32\\
JKNet-V2    & 2     & 512\\
SageNet     & 2     & 16\\
SageNet-V2  & 2     & 512\\
\bottomrule
\end{tabular}
}
\end{sc}
\end{small}
\end{center}
\end{table}


\end{document}